\documentclass[letterpaper]{article} % DO NOT CHANGE THIS
\usepackage{aaai2026}  % DO NOT CHANGE THIS
\usepackage{times}  % DO NOT CHANGE THIS
\usepackage{helvet}  % DO NOT CHANGE THIS
\usepackage{courier}  % DO NOT CHANGE THIS
\usepackage[hyphens]{url}  % DO NOT CHANGE THIS
\usepackage{graphicx} % DO NOT CHANGE THIS
\usepackage{natbib}  % DO NOT CHANGE THIS AND DO NOT ADD ANY OPTIONS TO IT
\usepackage{caption} % DO NOT CHANGE THIS AND DO NOT ADD ANY OPTIONS TO IT
\usepackage{algorithm}
\usepackage{algorithmic}

\usepackage{newfloat}
\usepackage{listings}
\DeclareCaptionStyle{ruled}{labelfont=normalfont,labelsep=colon,strut=off} % DO NOT CHANGE THIS
\floatstyle{ruled}
\newfloat{listing}{tb}{lst}{}
\floatname{listing}{Listing}
\title{Efficient and Effective In-context Demonstration Selection with Coreset}
\author{
    Zihua Wang\textsuperscript{\rm 1},
    Jiarui Wang\textsuperscript{\rm 1}, 
    Haiyang Xu\textsuperscript{\rm 2}, 
    Ming Yan\textsuperscript{\rm 2}, 
    Fei Huang\textsuperscript{\rm 2},
    Xu Yang\textsuperscript{\rm 1},
    Xiu-Shen Wei\textsuperscript{\rm 1}, 
    Siya Mi\textsuperscript{\rm 3, \rm 4}, 
    Yu Zhang\textsuperscript{\rm 1}\thanks{Yu Zhang is the corresponding author.}
}
\affiliations{
    \textsuperscript{\rm 1}School of Computer Science and
 Engineering and the Key Laboratory of New Generation Artificial Intelligence
 Technology and its Interdisciplinary Applications, Southeast University, Nanjing 210096, China.\\
      \textsuperscript{\rm 2}Tongyi Lab, Alibaba Group.\\
       \textsuperscript{\rm 3}School of Cyber Science and Engineering, Southeast University, Nanjing 211189, China.\\
               \textsuperscript{\rm 4}Purple Mountain Laboratories, Nanjing 210000, China.\\
     zhang\_yu@seu.edu.cn
}

\begin{document}

\maketitle

\begin{abstract}
In-context learning (ICL) has emerged as a powerful paradigm for Large Visual Language Models (LVLMs), enabling them to leverage a few examples directly from input contexts. However, the effectiveness of this approach is heavily reliant on the selection of demonstrations, a process that is NP-hard. Traditional strategies, including random, similarity-based sampling and infoscore-based sampling, often lead to inefficiencies or suboptimal performance, struggling to balance both efficiency and effectiveness in demonstration selection.
In this paper, we propose a novel demonstration selection framework named Coreset-based Dual Retrieval (CoDR).
We show that samples within a diverse subset achieve a higher expected mutual information.
To implement this, we introduce a cluster-pruning method to construct a diverse coreset that aligns more effectively with the query while maintaining diversity. Additionally, we develop a dual retrieval mechanism that enhances the selection process by achieving global demonstration selection while preserving efficiency.
Experimental results demonstrate that our method significantly improves the ICL performance compared to the existing strategies, providing a robust solution for effective and efficient demonstration selection.
\end{abstract}

\iffalse
\begin{links}
    \link{Code}{https://aaai.org/example/code}
    \link{Datasets}{https://aaai.org/example/datasets}
    \link{Extended version}{https://aaai.org/example/extended-version}
\end{links}
\fi
\section{Introduction}
In-context learning (ICL) is a groundbreaking approach that eliminates the need for conventional, data-intensive training methods. This innovative technique uses in-context demonstrations as prompts to enable few-shot learning, allowing models to perform tasks with minimal examples. Originally developed for Large Language Models (LLMs)~\cite{GPT3, liu2021makes}, ICL has recently gained traction in Large Vision-Language Models (LVLMs)~\cite{achiam2023gpt,awadalla2023openflamingo, laurencon2023obelics} as well. It has shown remarkable effectiveness across various domains, including image captioning (IC), and visual question answering (VQA) tasks, underscoring its flexibility and potential across modalities.

\begin{figure}[htbp]
    \centering
    \includegraphics[width=0.92\columnwidth]{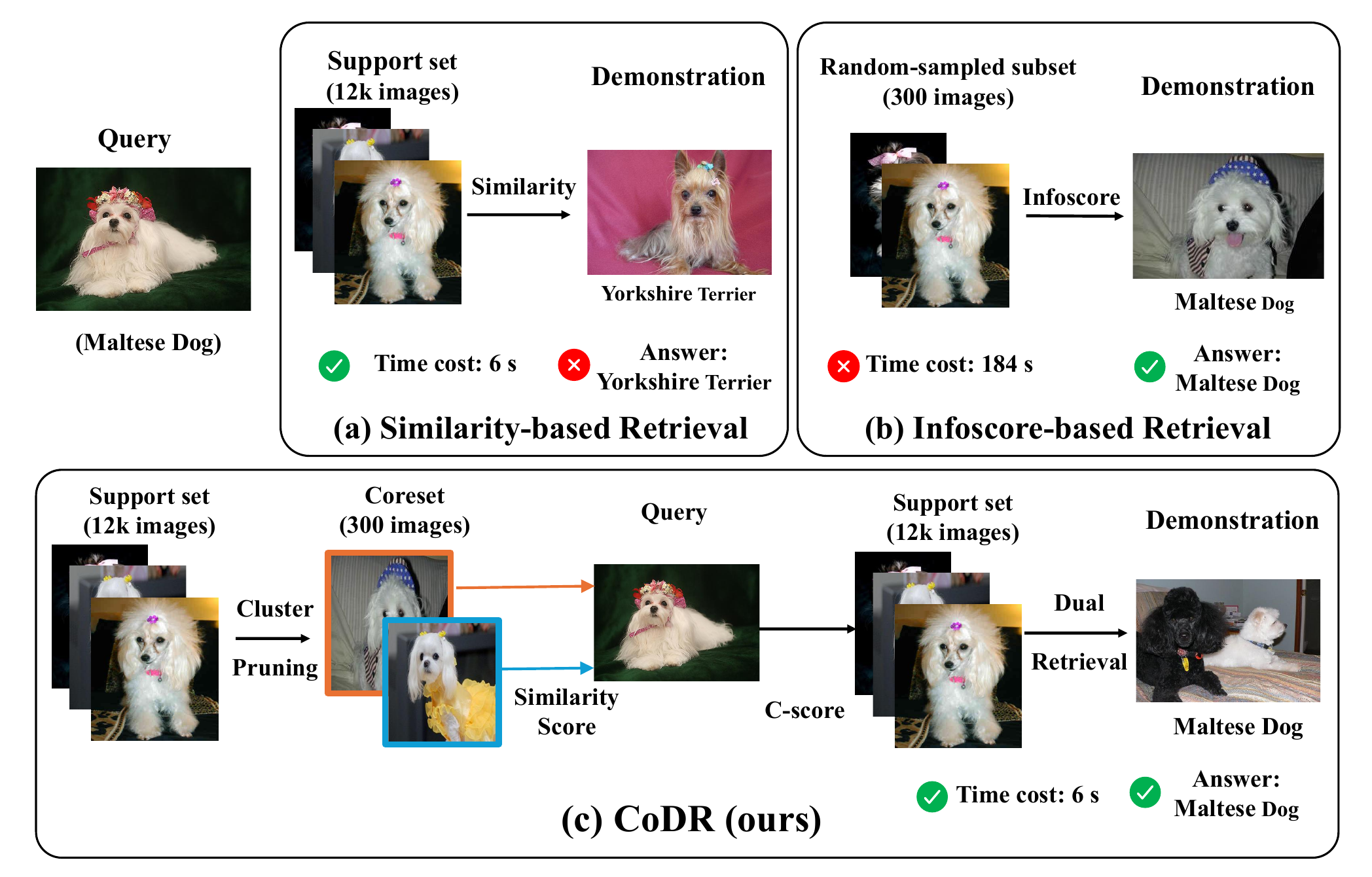}
    \caption{Similarity-based retrieval, infoscore-based retrieval and our proposed CoDR in ICL. CoDR introduces a dual retrieval mechanism: it first selects a coreset via cluster-pruning, then uses the similarity score between query and the samples in the coreset as a weighting coefficient, multiplying it with the pre-calculated C-score matrix. The weighted scores guide the final demonstration selection from the support set, achieving a more global selection.}
    \label{fig:infosimi}
\end{figure}

One of the primary obstacles is to select appropriate demonstrations according to different queries. This selection process is crucial, as the performance of LLMs and LVLMs heavily depends on the quality and relevance of the provided demonstrations~\cite{liu2021makes,laurencon2024matters,dong2022survey}.
However, selecting the appropriate demonstrations for each query is an NP-hard problem~\cite{li2023finding}. 
In other words, the quality of the selected demonstrations cannot be determined prior to the model execution of the inference process.
Existing selection strategies can be likened to a form of speculation in the selection process.
These strategies can be broadly categorized into two primary approaches. 
One focuses on efficiency, such as similarity-based selection~\cite{liu2021makes}.
However, similarity-based approaches are prone to hallucination, as the outputs tend to overly adhere to the input demonstrations instead of adequately addressing the query~\cite{liu2024survey}.
For example, as Figure~\ref{fig:infosimi} (a) shows, the image in the support set most similar to the query does not belong to the same dog breed, which is likely to lead to incorrect predictions when these samples are used as demonstrations. 
In instances where the model is presented with two highly similar images, and one is utilized as a demonstration, there is a propensity for the model to inaccurately classify the query image and the demonstration as belonging to the same category, despite this not being the case.
This issue is especially pronounced in tasks that require understanding of fine-grained image details, where superficial similarity fails to capture essential distinctions and leads to hallucinations~\cite{Yongliang_fcs}.

Another approach prioritizes effectiveness as the key property. For example, infoscore-based methods~\cite{li2023finding} focus on selecting demonstrations that maximize the mutual information gain between the query and the samples in the support set.
Infoscore is calculated by assessing each example’s influence on the predicted probabilities of others, a process that requires traversing the entire support set, making it time-consuming. Consequently, existing infoscore-based approaches often limit their demonstration search to a subset of the support set, typically sampled randomly from the full support set~\cite{li2023finding,yang2023lever}.
Moreover, unlike traditional image–text retrieval methods~\cite{cao2022image,cao2025multilingual,xu2025efficient}, the ground truth for the query is unavailable, making it infeasible to compute the actual infoscore value.
Instead, the sample with the highest probability is used to approximate the ground truth as an estimation (note that in the context of infoscore-based retrieval mentioned below, this refers to the estimated value).
As shown in Figure~\ref{fig:infosimi} (b), infoscore-based methods limit the demonstration selection to a subset of 300 images which are randomly sampled from the support set.
Despite this limitation, their computation time is significantly longer compared to similarity-based methods.
This constraint results in sub-optimal selections, which ultimately reduces the overall performance by not fully exploring the diversity of the entire support set.
Furthermore, the exponential growth of possible combinations further complicates the demonstration selection process, making it infeasible to enumerate all potential selections, especially as the number of demonstrations increases.

To address these challenges, we aim to develop a more effective and efficient demonstration selection strategy. First, to ensure the efficiency of the demonstration selection, we employ a subset of the support set, named coreset. 
Instead of being randomly sampled, the coreset is a subset that guarantees diversity.
We mathematically prove that such a coreset leads to higher mutual information, enabling better transfer of knowledge from the demonstrations.
To achieve diversity, we implement a cluster-pruning strategy to refine the supporting set into a coreset, where more complex clusters retain more samples after pruning. 
%This ensures that the coreset maintains the diversity needed for effective demonstration selection.

Due to the absence of query ground-truth label, it becomes challenging to directly and qualitatively assess which demonstrations in the support set are most effective.
However, the coreset, sampled from the support set, contains ground-truth labels. 
For each coreset sample, we treat it as a query and define the C-score to evaluate how well each sample in the support set fits the query.
As Figure~\ref{fig:infosimi} (c) shows, for a given query, we first measure its similarity with each sample in the coreset.
We then multiply this similarity score by the C-score of each coreset sample across the support set to obtain the accumulated scores. 
These accumulated scores provide a measure of the quality of each sample in the support set.
Based on these scores, we select the top-k samples to form the k-shot demonstrations for the given query.
As a result, it enhances performance by ensuring that the selected demonstrations are not limited to the coreset, thus having a broader and more effective selections while simultaneously maintaining efficiency.
Compared to the similarity-based retrieval method, CoDR has almost the same retrieval time, but shows improvements of $7.01$/$5.18$/$5.78$ on the IC\_CIDEr/VQA\_ACC/FIC\_ACC metrics, respectively.
When compared with the infoscore-based method, CoDR demonstrates performance advantages of $2.40$/$1.34$/$1.79$, with nearly $4$ times faster retrieval speed.

Our contributions are concluded as follows:

\begin{itemize}
\item We demonstrate that a more diverse support set leads to higher mutual information expectation, enabling more effective knowledge transfer from the demonstrations.
\item We introduce a coreset-based demonstration selection strategy that effectively balances diversity and relevance in visual-language in-context demonstration selection.
\item Experimental results on image captioning, visual question answering, and fine-grained image classification tasks demonstrate that our approach significantly improves the performance of visual-text ICL on both the OpenFlamingo-v2 and Idefics-v2 models.
\end{itemize}

\section{Related Works}
\subsection{Models with ICL Ability}
ICL has emerged as a transformative paradigm in machine learning, with the potential to replace traditional few-shot training approaches by utilizing a few examples as the input~\cite{mosbach2023few}.
Powered by LLMs like GPT-3~\cite{GPT3}, LLaMA~\cite{dubey2024llama}, and MPT~\cite{team2023introducing}, ICL has found widespread application in textual generation tasks, \textit{e.g.}, classification~\cite{edwards2024language}, question answering~\cite{venktesh2023context,peng2024live}, table processing~\cite{lu2025large}, and reading comprehension~\cite{li2023evaluation}.
Similarly, LVLMs like Flamingo~\cite{alayrac2022flamingo}, Idefics~\cite{laurencon2023obelics,laurencon2024matters}, GPT4~\cite{achiam2023gpt}, leverage ICL to integrate image and text information, enabling richer Visual-Language (VL) tasks, \textit{e.g.}, visual question answering~\cite{li2024configure, nie2024code}, image captioning~\cite{yang2023lever,yang2024exploring,Zheng_fcs}, image classification~\cite{zhang2024instruct}, and segmentation~\cite{wang2023seggpt, Wen_2025_ICCV}.
Among the LVLMs with ICL capabilities, we employ OpenFlamingo-v2~\cite{awadalla2023openflamingo} and Idefics-v2~\cite{laurencon2023obelics} due to their robust ICL performance and open-source availability. This transparency ensures that neither model has been trained on the query or support sets, reducing potential bias.

\subsection{ICL Demonstration Strategy}
To enhance the performance of ICL, researchers have explored a variety of strategies, primarily centered on optimizing the quality and structure of demonstrations. Early approaches emphasized improving demonstrations through formatting adjustments~\cite{lu2021fantastically} or reordering ~\cite{kumar2021reordering}, rather than selective curation of instances.
Recent research has increasingly emphasized selecting high-quality demonstrations to further optimize model performance.
For example, metrics such as perplexity~\cite{qin2023context}, BERTScore recall~\cite{gupta2023coverage}, and mutual information~\cite{li2023finding, liao2022zero} are regarded as useful indicators of demonstration quality. 
However, some studies suggest that these metrics may be effective only for specific tasks~\cite{wu2022self, li2023unified}.
Demonstration selection strategies have also been found to be both data- and model-dependent~\cite{peng2024revisiting,zhu2025exploring}.
Recent studies~\cite{guo2023images,feng2024auto} have explored the use of LLM-generated demonstrations to support ICL. While effective in some scenarios, these learned retrievers~\cite{feng2024auto,shen2024retrieval,askari2025magic} invoke LLMs for each query, leading to high computational overhead. Moreover, the quality of the retrieved demonstrations is highly sensitive to the initial prompt.  
Some methods employ calibration techniques, such as iterative refinement~\cite{qin2023context}, while others enhance both the prompt and the demonstration~\cite{yao2024more,jiang2025mimic}.
A common limitation of these methods is their reliance on subsampling, as demonstrations are typically selected from a randomly drawn subset of the full support set, rather than the entire dataset. This constraint often arises due to computational or scalability challenges, yet it may lead to sub-optimal results. 

\begin{figure*}[t]
    \centering
    \includegraphics[width=1\textwidth]{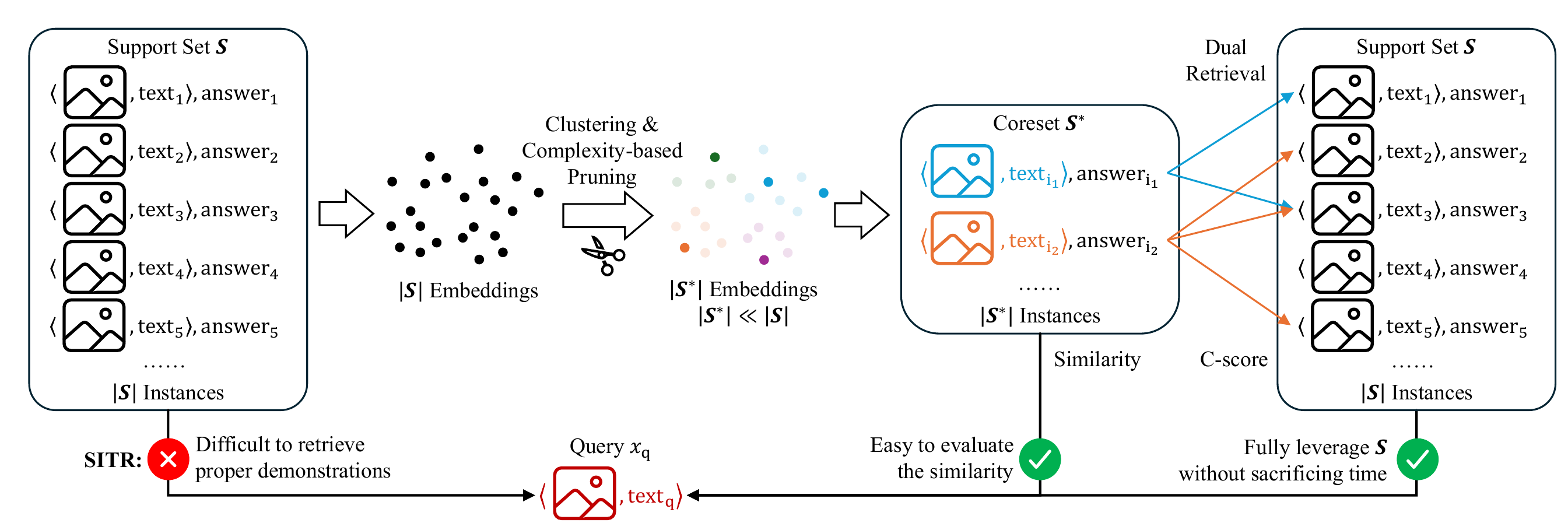}
    \caption{Architecture of CoDR. Our method construct a coreset {\boldmath$S^*$} by clustering and pruning the support set {\boldmath$S$}. A dual retrieval module then performs global retrieval as follows: We precompute a C-score quantifying each support set sample's quality as a demonstration when queried with coreset samples. For an input query, we evaluate its similarity to each coreset sample, hypothesizing that higher similarity correlates with more similar demonstration selection. The final demonstration score is the product of this similarity and the precomputed C-score, enabling global retrieval across the entire support set.}
    \label{fig: archi}
\end{figure*}

\section{Coreset-based Dual Retrieval (CoDR)} 
%xxx, overall
In this section, we begin by mathematically demonstrating that a diverse coreset yields a higher expectation of mutual information. Subsequently, we describe a method for constructing a coreset with sufficient diversity. Lastly, we introduce dual retrieval, which leverages this diverse coreset to efficiently identify effective  demonstrations from the entire support set using the C-score. The framework of CoDR is shown in Fig.~\ref{fig: archi}.

\subsection{Preliminaries} 
Given a $n$-shot VL demonstrations ${\langle x_i, y_i \rangle}_{i=1}^n$, where $x_i$ represents the multimodal input (\textit{e.g.}, an image and its associated question in VQA), and $y_i$ is the corresponding label (\textit{e.g.}, the answer), along with a query input $x_{q}$,
They are concatenated as $x' =\mathrm{concat}(\langle x_i,y_i \rangle _{i=1}^n, x_{q})$ and fed into the LVLM, which then generates the predicted output $\hat y_{q}$.

\subsection{Coreset Construction} 
\label{sec:coreset}
For a specific query $x_{q}$, selecting an appropriate set of demonstrations from the support set {\boldmath$S$} is crucial to the ICL performance, yet remains a challenging task.
Although existing strategies, such as similarity-based~\cite{zebaze2024context,an2023context} and infoscore-based approaches~\cite{li2023finding,yang2023lever}, offer practical heuristics, they often suffer from inefficiency in large-scale support sets and limited effectiveness in capturing semantic diversity relevant to the query.
Our objective is to identify a compact subset {\boldmath$S^*$} $\subset$ {\boldmath$S$}  that captures the most informative demonstrations for inference. We refer to this subset {\boldmath$S^*$} as the coreset.
Intuitively, a diverse subset offers broader coverage of the semantic space, potentially reducing the model’s uncertainty when making predictions. To formalize this intuition, we employ mutual information as a metric to quantify the informativeness of the selected subset with respect to the query's target output.
Next, we provide a theoretical justification for this assumption and introduce a practical method for constructing the coreset.

We define the mutual information to evaluate the input $x'$ and the model’s target prediction $y_q$ as:

\begin{equation}
    I(x', y_q) = H(y_q) - H(y_q \mid x'),
\label{eq:MI}
\end{equation}
where $H$ represents the entropy.
Since $H(y_q)$ is independent of the chosen demonstrations (assuming the same query $x_q$), maximizing $I(x', y_q)$ reduces to minimizing the conditional entropy $H(y_q \mid x')$.

We consider two subset selection scenarios: {\boldmath$S$}$_\mathrm{div}$, a diverse set of demonstrations sampled to maximize semantic variance, and {\boldmath$S$}$_\mathrm{rand}$, a randomly sampled set of demonstrations.
To evaluate the quality of the selected subsets, we compare the expected conditional entropy between the query and the predicted label given the combined input: ${E}[H(x'_\mathrm{div}, y_q)]$ versus ${E}[H(x'_\mathrm{rand})]$, where $x'_\mathrm{div}$ and $x'_\mathrm{rand}$ denote the concatenation of the query $x_q$ with demonstrations from the {\boldmath$S$}$_\mathrm{div}$ and {\boldmath$S$}$_\mathrm{rand}$, respectively.
The key idea is that the mutual information between a query and the support set increases when the support set provides better coverage of the query’s latent class. 
To operationalize this notion of class coverage, we approximate the latent class of each sample.
Since most of the VL tasks like VQA and image captioning lack category labels, we ``assigned'' categories (\textit{e.g.}, through clustering the latent features).

Therefore, the conditional entropy (the second term of the Eq.~\ref{eq:MI}) can be rewritten as:
\begin{equation}
     H(y_q \mid x')= -\sum P(y_q \mid x') \log P(y_q \mid x').
\end{equation}
Since the function `` $\cdot \log (\cdot)$'' is convex, a diverse set—by evenly spreading probability mass across latent categories—leads to lower expected conditional entropy via Jensen’s inequality. This directly implies that:
\begin{equation}
E[H(y_q \mid x'_{\mathrm{rand}})] = H\left[y_q \mid E(x'_{\mathrm{rand}})\right] \geq E[H(y_q \mid x'_{\mathrm{div}})],
  \label{eq:entropy}
\end{equation}
where $E$ represents the mathematical expectation.
Consequently, the subset with diverse demonstrations is expected to yield greater mutual information:
\begin{equation}
{E}[I(x'_\mathrm{div},y_q)] \geq {E}[I(x'_\mathrm{rand},y_q)].
\end{equation}

While applying greedy algorithms or K-means clustering can yield diverse subsets from the support set, these methods often fall short in ICL.
This is because they prioritize more diverse sample features at the expense of considering the relevance and alignment between the demonstration samples and the query input. Thus, a more balanced approach that considers both diversity and relevance is essential to improve performance in ICL.

With the guidance of mutual information theory, we employ a cluster-based pruning method to obtain a subset {\boldmath$S^*$} of the support set {\boldmath$S$}.
We notice that Density-Based Pruning (DBP) and self-supervised pruning (SSP-Pruning) are effective in pruning the training set in image classification~\cite{kamal2024density,NEURIPS2022}.
They prune the cluster items based on similarity or diversity.
Inspired by them, we use the LVLM to obtain the embedding of the support set {\boldmath$S$}, and then perform K-means clustering on this feature.
However, we propose that pruning should be based on complexity $F$, which is defined as follows:
\begin{equation}
    F_i =  \overline{d(l,m)}_{l,m \in K_i} \cdot   \overline{d(l,m)}_{l\in K_i,\\ m\notin K_i},
\end{equation}
where $d$ represent the cosine distance between features. The first term of the formula denotes the intra-cluster distance for cluster $K_i$, measuring the average distance among points within the same cluster.
The second term represents the average inter-cluster distance, calculating the mean distance between the points in cluster $K_i$ and the points in other clusters.
Intuitively, a cluster with high intra-cluster distance requires more retained samples to preserve its internal variability. Meanwhile, high inter-cluster distance indicates uniqueness, necessitating sufficient representation to avoid losing distinct features. This dual perspective ensures that pruning retains informative clusters while reducing redundancy in over-represented ones.
For clusters with higher complexity, more samples should be retained after pruning.
Therefore, we preserve $\lfloor |K_i| \cdot \exp(F_i)/\sum_{j=1}^{n} \exp(F_j) \rfloor$ samples in cluster $K_i$.
Since the coefficient $\exp(F_i)/\sum_{j=1}^{n} \exp(F_j)$ is strictly less than 1, this ensures the pruning process.
We obtain the coreset {\boldmath$S^*$} after cluster-pruning.

\subsection{Dual Retrieval} 
\label{sec:cscore}

After obtaining the coreset {\boldmath$S^*$}, which ensures diversity, we note that due to the reduced sample size compared to the entire support set {\boldmath$S$} after multiple pruning steps, selecting samples exclusively from the coreset may not yield a globally optimal solution.
While the coreset effectively captures the most representative samples from the support set, it is limited in its ability to fully leverage the variety of available demonstrations on larger datasets. 
Therefore, we aim to use the coreset as a guide to retrieve samples globally.

We define a C-score $C(q|s)$ to quantify how well a demonstration $s \in$ {\boldmath $S$} supports the query $q = \langle x_q,y_q \rangle$.
\begin{equation}
    C(q|s)=\mathrm{Metric}(\mathrm{LVLM}(\mathrm{concat}(s, x_{q})), y_{q}),
\end{equation}
where $\mathrm{Metric}(\hat y_q,y_q)$ is a metric function determined by the task, with larger values indicating more accurate predictions $\hat{y}_q$ (e.g., CIDEr for image captioning and ACC for VQA). However, with $q \in$ {\boldmath$S_{q}$}, $y_q$ is unknown during inference.

Given the premise of ICL, where $x_{q}$ and $x$ belong to the same domain, we assume that {\boldmath$S_{q}$} (query set) and {\boldmath $S$} (support set) follow the same distribution.
While {\boldmath$S_{q}$} is unannotated, {\boldmath $S$} contains ground-truth labels, allowing us to leverage labeled samples from {\boldmath $S$} for ICL purposes.
In this way, $C(q^*|s)$ can be easily obtained, where $s \in$ {\boldmath $S$} and $q^* \in$ {\boldmath $S^*$}.
Due to the characteristics of the clustering, the retained samples in {\boldmath $S^*$} ensure diversity.
Therefore, the C-score $C(q^*|s)$ can be regarded as kernels to represent the samples in {\boldmath$S_{q}$}:
\begin{equation}
    \hat C(q|s)=\frac{\sum_{q^*\in \textbf{\boldmath$S^*$}} [d(q, q^*)\cdot C(q^*|s)]}{\sum_{q^*\in \textbf{\boldmath$S^*$}} d(q, q^*)}.
\end{equation}
Based on the cosine similarity between the query sample $q$ and the coreset samples in {\boldmath$S^*$}, the contribution of each kernel can be calculated and thus obtain the score $\hat C(q|s)$.
This process generates a score set indicating how effectively each demonstration $s \in$ {\boldmath $S$} supports the query sample $q \in$ {\boldmath$S_{q}$}.
Importantly, the C-score matrix $C (q^*| s) $ for each coreset sample $q$ can be precomputed once per task and reused across queries. 
Unlike traditional similarity-based retrieval methods, which compute pairwise similarities across the entire support set, our approach merely calculates similarities between the input query and the coreset, which enables efficient large-scale deployment on demonstration selection.

From the perspective of ensemble learning, each 1-shot demonstration can be regarded as a weak learner. Aggregating multiple demonstrations is akin to combining weak learners to enhance overall performance. In the $n$-shot setting, selecting diverse and high-quality examples provides complementary information, helping to reduce the risk of overfitting to any single instance. Therefore, we construct the $n$-shot prompt by selecting the top-$n$ demonstrations based on their relevance scores, ensuring that the final representation is both informative and reliable for the query.

\section{Experiments}

\subsection{Tasks and Datasets}
Our approach is evaluated on 3 tasks: Image Captioning (IC), Visual Question Answering (VQA), and Fine-grained Image Classification (FIC).
We evaluate on two mainstream open-source LVLM frameworks: OpenFlamingo-v2 (OFv2)~\cite{awadalla2023openflamingo} and Idefics-v2 (IDEv2)~\cite{laurencon2023obelics}.
We employ MSCOCO~\cite{lin2014microsoft}, VQAv2~\cite{antol2015vqa}, StanfordDogs~\cite{StanfordDogs} for IC, VQA, FIC tasks, respectively.

\subsection{Implementation Details}

%\hspace{\parindent}
%\textbf{Shot number $n$.} 
\paragraph{Shots number $n$.}
Following the principle of diminishing marginal returns, our evaluation primarily focuses on $1$-/$2$-/$4$-shot.
Our ablation studies are conducted using a $4$-shot configuration.
%Additionally, results for $8$-shot scenarios are provided in the Supplementary materials.

\paragraph{Implementing ICL.}
We set the maximum number of generated tokens to $16/5/5$ for IC/VQA/FIC tasks, respectively. For fluent sentence generation, we employ beam search with a beam size of $3$. 
The features used for clustering and similarity-based retrieval are extracted by the encoder of the LVLMs (OFv2 and IDEv2).
The inference process is conducted on an Nvidia A6000 GPU.

\paragraph{Metrics.}
To evaluate the effectiveness of VQA and FIC, we calculate accuracy scores (VQA\_ACC and FIC\_ACC), with higher scores indicating better performance of the VQA model. For the IC task, we observe that while the sentences generated through ICL may yield high scores, they can also contain factual errors. Therefore, in addition to traditional metrics CIDEr~\cite{vedantam2015cider}, we further evaluate hallucination metrics, CHAIRs and CHAIRi~\cite{rohrbach2018object}, as hallucinations are considered one of the severe challenges in LLM generations~\cite{huang2023survey}.

\begin{table*}[ht]
\centering

\begin{tabular}{lcccccccccc}
\hline
Model-Method & \multicolumn{3}{c}{CIDEr $\uparrow$} & \multicolumn{3}{c}{CHAIRs $\downarrow$} & \multicolumn{3}{c}{CHAIRi $\downarrow$} & Time(s)$\downarrow$ \\
\hline
& 1-shot & 2-shot & 4-shot & 1-shot & 2-shot & 4-shot & 1-shot & 2-shot & 4-shot & \\
\hline
OFv2-RG                       & 74.33          & 85.65           & 95.20           & 8.8          & 6.5          & 6.3          & 7.9          & 6.3          & 6.3          & -                    \\
OFv2-RL                       & 74.23          & 85.40           & 95.14           & 8.5          & 6.5          & 6.2          & 7.9          & 6.3          & 6.3          & -                    \\
OFv2-SITR                     & 73.10          & 82.99           & 97.29           & 19.32        & 8.3          & 6.2          & 14.35        & 6.6          & 5.9          & 5.37                 \\
OFv2-SITRL                    & 73.40          & 82.94           & 97.22           & 17.10        & 8.9          & 6.2          & 13.1         & 6.7          & 6.0          & \textbf{3.81}        \\
OFv2-IRL                      & 74.95          & 89.06           & 101.83          & 5.9          & 5.2          & 4.3          & 5.4          & 3.9          & 3.7          & 13.78                \\
OFv2-CoDR                     & \textbf{79.68} & \textbf{100.80} & \textbf{104.23} & \textbf{5.9} & \textbf{5.1} & \textbf{4.2} & \textbf{5.3} & \textbf{3.4} & \textbf{3.3} & 3.95                 \\ \hline
IDEv2-RG                            & 75.59          & 100.33          & 112.70          & 7.4          & 4.9          & 5.2          & 7.3          & 3.9          & 3.9          & -                    \\
IDEv2-RL                            & 75.43          & 99.67           & 109.30          & 7.4          & 5.0          & 5.2          & 7.3          & 3.9          & 3.9          & -                    \\
IDEv2-SITR                          & 74.55          & 92.91           & 102.71          & 10.7         & 6.9          & 6.6          & 8.2          & 5.0          & 4.8          & 4.85                 \\
IDEv2-SITRL                         & 74.59          & 92.70           & 101.76          & 13.9         & 6.8          & 5.3          & 7.9          & 5.0          & 4.5          & \textbf{3.74}        \\
IDEv2-IRL                           & 79.61          & 105.01          & 111.29          & 5.6          & 5.2          & 4.8          & 5.6          & 5.1          & 4.6          & 13.35                \\
IDEv2-CoDR                  & \textbf{89.57} & \textbf{115.06} & \textbf{117.54} & \textbf{4.9} & \textbf{3.7} & \textbf{4.3} & \textbf{3.6} & \textbf{2.9} & \textbf{3.1} & 3.77                 \\ \hline
\end{tabular}
\caption{1-/2-/4-shot IC performance with various demonstration selection strategies.}
\label{tab:result_cap}
\end{table*}

\begin{table}[htbp]
\centering

\begin{tabular}{lcc}
\hline
Model-Method & VQA\_ACC $\uparrow$ & FIC\_ACC $\uparrow$ \\
\hline
{OFv2}-RG     &   41.97/45.92/48.95      &   17.83/28.51/34.36    \\
{OFv2}-RL     &   42.12/46.07/48.99       &  18.56/29.29/33.16         \\
{OFv2}-SITR   &   40.17/43.58/47.50      &      29.33/30.25/36.98    \\
{OFv2}-SITRL  &   42.09/44.72/48.31       &     27.13/32.84/35.61     \\
{OFv2}-IRL    &   46.66/50.83/52.15   &        29.88/33.79/39.60  \\
{OFv2}-CoDR    &   \textbf{50.03}/\textbf{52.53}/\textbf{53.49}       &   \textbf{33.43/39.38/41.39 }        \\ \hline
{IDEv2}-RG     &   60.24/63.20/66.62       & 43.94/43.92/44.85        \\
{IDEv2}-RL     &    60.03/64.23/66.71      &     41.70/41.97/43.02     \\
{IDEv2}-SITR   &    58.65/62.50/66.93      &     47.14/53.10/54.42     \\
{IDEv2}-SITRL  &    59.17/62.20/66.78      &     47.22/52.89/54.00     \\
{IDEv2}-IRL    &    60.65/64.46/\textbf{67.04}    &     43.85/49.88/50.17     \\
{IDEv2}-CoDR    &    \textbf{61.16}/\textbf{64.53}/67.01      &     \textbf{52.74}/\textbf{58.64}/\textbf{66.28}     \\ \hline
\end{tabular}
\caption{1-/2-/4-shot VQA and FIC performance with various demonstration selection strategies.}
\label{tab:result_vqa_fic}
\end{table}

\subsection{Results and Analyses}
\subsubsection{Methods for Comparison.}
%We compare our method to 5 main selection strategies:

\noindent(1) \textbf{Random-Global (RG)}: RG randomly samples demonstrations from the whole support set {\boldmath$S$}.
(2) \textbf{Random-Local (RL)}: RL first randomly samples a subset of the support set {\boldmath$S$}, then the demonstrations are randomly selected from this subset.
(3) \textbf{Similarity-based Image-Text Retrieval-global (SITR)}: Benefiting from the capabilities of CLIP, the similarity between the features of image-text demonstrations and the query can be aligned in the feature domain. Searching for demonstrations in support set $S$ that are more similar to the query has been recognized as a potentially useful metric.
(4) \textbf{Similarity-based Image-Text Retrieval-Local (SITRL)}: Similar to the RL strategy, we first randomly sample a subset of the support set and then conduct a search based on image-text similarity within this subset.
(5) \textbf{Infoscore-based Retrieval-Local (IRL)}: Infoscore~\cite{yang2023lever,li2023finding} can be considered a greedy search strategy due to its high complexity and unsuitability for global search. Consequently, we utilize Infoscore to search for demonstrations within a randomly selected subset of 300 samples.

\subsubsection{Main Results.}
The results of the various demonstration selecting strategies are listed in Table~\ref{tab:result_cap} (IC) and Table~\ref{tab:result_vqa_fic} (VQA and FIC).
In IC task using the OFv2 model, our method significantly enhances the CIDEr while reducing the CHAIRi and CHAIRs compared to the random algorithms RG and RL.
It outperforms both SITR and SITRL methods, particularly excelling in the CHAIR metrics, which indicates that CoDR reduces the hallucinations.
In the IDEv2 model, we observe a similar trend, with our method attaining a CIDEr score of $89.57/115.06/117.54$ in the $1$-/$2$-/$4$-shot scenarios. While other methods like SITR and IRL struggle to outperform RG on IDEv2 (likely due to the model’s inherent complexity or feature distribution), CoDR’s consistent performance highlights its robustness across different model architectures. This indicates that CoDR effectively balances relevance and diversity.

Table~\ref{tab:result_cap} also lists the average inference time required to process each query. Our method achieves average processing times of $3.95$ seconds (OFv2) and $3.77$ seconds (IDEv2), closely matching SITRL’s performance ($3.81$ and $3.74$ seconds, respectively). Compared to SITR, which is also a global search method, our method demonstrates a 1.42 second speed improvement for OFv2 and a 1.08 second improvement for IDEv2. These results demonstrate that CoDR achieves significantly better computational efficiency than conventional similarity-based retrieval methods.

Our method achieves near state-of-the-art performance in both VQA and FIC tasks compared to other strategies.
In the IDEv2 4-shot VQA setting, CoDR attains an accuracy of $67.01$, closely matching IDEv2-IRL ($67.04$), with a marginal difference of only $0.03$. 
For the OFv2 model, our method demonstrates a significant improvement, achieving the highest VQA accuracy of $50.03$/$52.53$/$53.49$ under the $1$-/$2$-/$4$-shot setting, compared to other strategies such as SITR and IRL, which yield accuracies of $40.17/43.58/47.50$ and $46.66/50.83/52.15$, respectively. In the FIC task, our approach outperforms the baseline methods, reaching $33.43/39.38/41.39$ for the $1$-/$2$-/$4$-shot setting.
When using the IDEv2 model, the performance advantage in VQA for our method is relatively smaller compared to FIC. Similarly, other approaches such as SITR and IRL do not significantly outperform RG, likely due to limitations of the IDEv2 architecture.  Nevertheless, CoDR maintains the strongest overall performance.

The performance of RG and RL is relatively close because the subset selected by the RL strategy is randomly sampled from the support set, which leads to a distribution that is expected to be similar to that of the support set, resulting in comparable performance.
This similar pattern is also observed in SITR and SITRL. Furthermore, we find that similarity-based methods do not outperform random sampling in VQA, whereas they remain a relatively effective strategy in FIC.
In the IC task, while similarity-based methods yield higher CIDEr scores, they also introduce excessive hallucinations, resulting in poorer CHAIRi and CHAIRs metrics. This occurs because the model tends to focus on the similar demonstrations while neglecting the query sample.
In FIC, finding the most similar image may lead to selecting images from the same category, which are then adopted as the output by the model. However, due to the differences in labeling information between VQA and IC, this approach can inadvertently introduce hallucinations.
In contrast, our method employs clustering to ensure the diversity of samples during the coreset selection process. Subsequently, the dual retrieval process effectively identifies the appropriate demonstrations based on the input query, thereby enhancing the performance of IC, VQA, and FIC tasks.

\subsubsection{Ablations}

\begin{table}[htbp]

\centering
\begin{tabular}{lcccc}
\hline
Model-Task & Rand & Centroids& \textbf{CoDR}  \\
\hline
OFv2-IC    & 80.12    &  101.09   & \textbf{104.23} \\
OFv2-VQA   & 46.86    &  52.24   & \textbf{53.49}  \\
OFv2-FIC   & 33.64    &  40.97   & \textbf{41.39}  \\
IDEv2-IC   & 113.93   &  117.22  & \textbf{117.54} \\
IDEv2-VQA  & 64.37    &  66.38   & \textbf{67.01}  \\
IDEv2-FIC  & 50.25    &  63.56   & \textbf{66.28}  \\
\hline
\end{tabular}
\caption{Ablations on different coreset selection methods.}
\label{tab:ablation_cp}
\end{table}

\begin{table}[htbp]

\centering
\begin{tabular}{lcccc}
\hline
Model-Task & Rand & Simi & Div & \textbf{CoDR}  \\
\hline
OFv2-IC    & 78.30         & 97.83   &  79.05    & \textbf{104.23} \\
OFv2-VQA   & 43.58         & 44.18   &  45.83    & \textbf{53.49}  \\
OFv2-FIC   & 29.53         & 36.47    &  24.44   & \textbf{41.39}  \\
IDEv2-IC   & 117.20        & 113.20   &  115.13   & \textbf{117.54} \\
IDEv2-VQA  & 66.83         & 66.33    &  66.80  & \textbf{67.01}  \\
IDEv2-FIC  & 42.09         & 52.50    &  49.28   & \textbf{66.28}  \\
\hline
\end{tabular}
\caption{Ablations on the effectiveness of dual retrieval.}
\label{tab:ablation_dr}
\end{table}

%xxxxxx
\begin{figure*}[htbp]
    \centering
    \includegraphics[width=2\columnwidth]{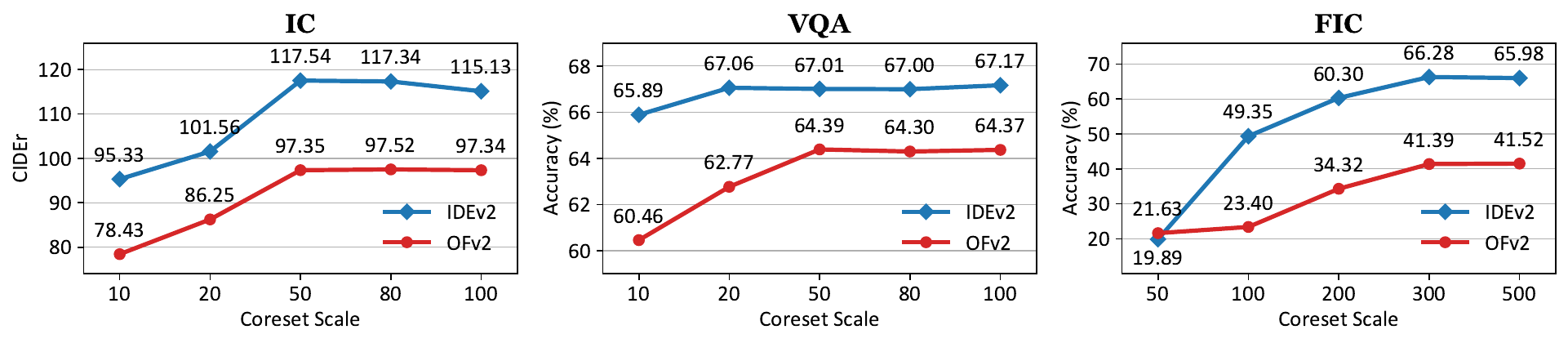} % Single image filling one column
    \caption{IC, VQA and FIC performance with different coreset scales.}
    \label{fig:ablation}
\end{figure*}

%xxxxxx
\paragraph{Coreset construction.}  We evaluate the quality of coreset obtained using different methods. As shown in Table~\ref{tab:ablation_cp}, we compare three approaches: random sampling (Rand), selecting samples closest to the cluster centroids (Centroids), and our proposed cluster-pruning method (CoDR). 
The results demonstrate that CoDR consistently achieves the best performance, which indicates that CoDR is effective in constructing a more representative and informative coreset.
Moreover, CoDR also consistently outperforms the cluster-center baseline, which uses the centroids directly as coreset samples. This method captures coarse semantic structure, but fails to ensure optimal representativeness or diversity. CoDR, by contrast, applies additional pruning or normalization to select more globally informative samples. For example, CoDR improves over the cluster-center method by +3.14 on OFv2-IC, +2.72 on IDEv2-FIC, and +1.13 on OFv2-VQA, showing that fine-grained coreset selection beyond simple clustering is crucial for performance.

\begin{figure}[htbp]
\centering
    \includegraphics[width=\columnwidth]{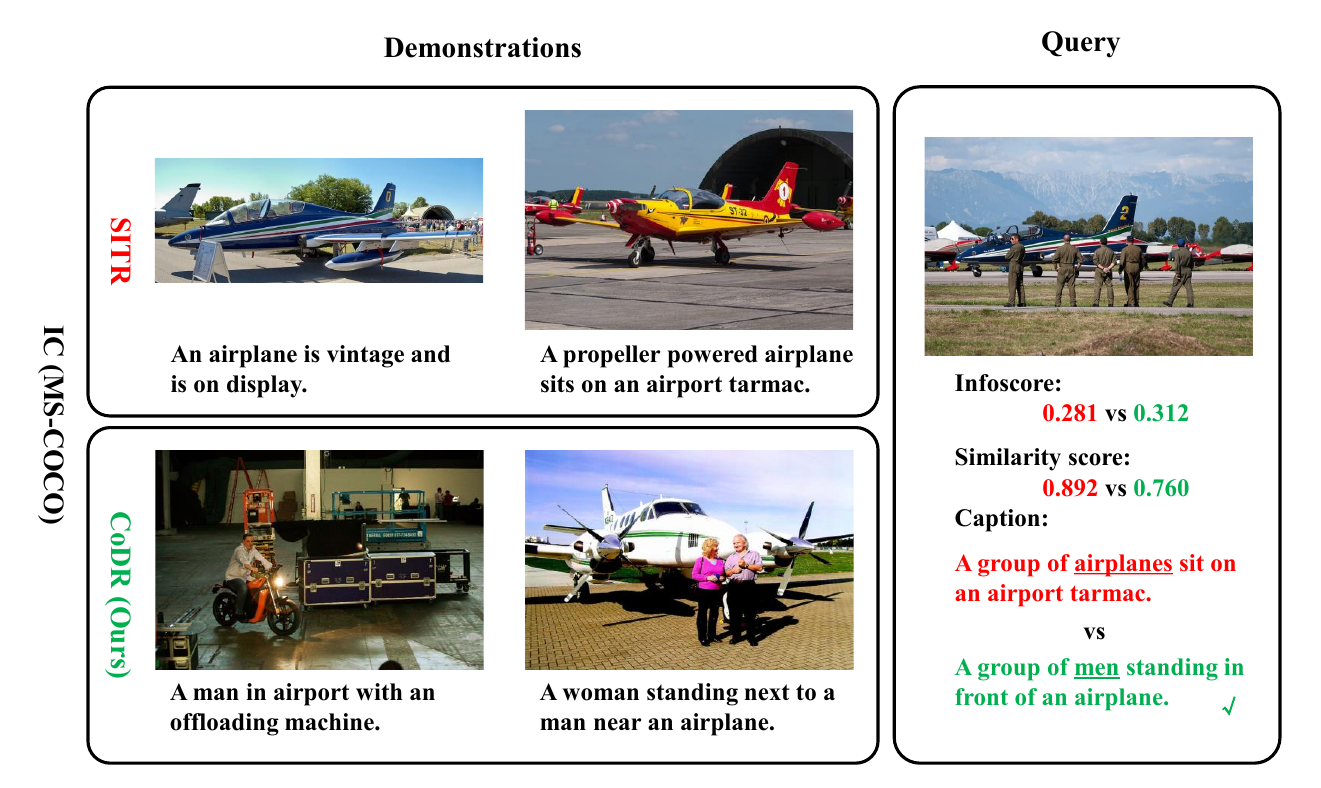}
    \caption{Visualization of SITR and CoDR on IC.}
    \label{fig:vis}
\end{figure}

\paragraph{Coreset Scale.}
The coreset size serves as the primary hyperparameter in the cluster-pruning process. We systematically investigate the performance of IC, VQA, and FIC under varying coreset scales. While larger coreset sizes expand the candidate sample pool for selection, this expansion also increases the probability of encountering redundant samples with high similarity, consequently elevating the computational overhead required to identify optimal demonstrations.
The results in Fig.~\ref{fig:ablation} indicate that a preferred coreset scale is approximately 50 for IC, 50 for VQA, and 300 for FIC.
Due to the absence of category labeling for image-text pairs in VQA and IC tasks, a coreset scale of 50 samples is sufficient for effective sample selection. In contrast, FIC task has 120 categories (for the Stanford Dogs dataset), and when the coreset scale is less than 120, certain categories are absent, leading to performance degradation. However, with a coreset scale above 300, there is a higher likelihood that all 120 categories are represented in the coreset at least once, which indeed occurs in practice.
%More ablation results on clustering sensitivity (including multi-stage clustering and random seeds) are listed in Supplementary materials.

\paragraph{Dual Retrieval.} 

We evaluate the effect of dual retrieval by comparing it with direct selection from the coreset using three alternatives: random (Rand), similarity-based (Simi), and diversity-based (Div) sampling, with results presented in Table~\ref{tab:ablation_dr}.
As shown in the table, the dual retrieval approach consistently outperforms random selection (Rand) from the coreset, as it adopts a more global selection strategy that provides richer candidate examples, whereas the coreset itself remains relatively small.
We further observe that similarity-based sampling within the coreset generally surpasses random selection, since the coreset construction already ensures a certain degree of diversity.
In addition, random sampling from the coreset yields better results than the RG strategy, which randomly selects from the full support set, and the RL strategy, which samples from a randomly chosen subset of it.
Finally, diversity-based sampling consistently underperforms CoDR, particularly on OFv2-FIC, which demands strong semantic alignment. This indicates that maximizing diversity alone may introduce irrelevant examples, highlighting the necessity of jointly considering both relevance and diversity as in CoDR.

\subsection{Qualitative Results}
Fig.~\ref{fig:vis} visualizes a representative query-demonstration pair, highlighting CoDR's advantages over SITR. Unlike SITR, CoDR selects diversified demonstrations that reduce hallucination while preserving relevance. For example, SITR erroneously predicts 'airplanes' instead of 'men', a clear error propagation from its second demonstration that deviates from ground-truth. This misalignment reveals SITR's over-reliance on demonstration content, undermining query fidelity. In contrast, CoDR yields more faithful predictions with more diverse and representative demonstrations.
%More examples are present in Supplementary Materials.

\section{Conclusion}
In this paper, we propose a novel demonstration selection approach, named CoDR.
CoDR first prunes the support set into a compact yet diverse coreset, ensuring representational coverage across different data modes. With the guidance of mutual information, this diverse coreset facilitates richer information transfer from varied demonstrations, thereby improving generalization.
Subsequently, our dual retrieval mechanism leverages this coreset as a guide to efficiently identify task-relevant examples from the full support set.
Experimental evaluations validate the effectiveness of our approach, showing significant performance improvements over traditional selection strategies.
In the future, we plan to extend CoDR to knowledge-intensive tasks, where identifying demonstrations that encode both factual and relational knowledge will be crucial in complex multimodal scenarios.

\subsubsection{Acknowledgments.}
This work was supported by National Key R\&D Program of China (2021YFA1001100), National Natural Science Foundation of China under Grant (62576089, 62522602, 62576091), and the Fundamental Research Funds for the Central Universities (2242025K30024, 4009002401). This research work is supported by the Big Data Computing Center of Southeast University.

\begin{small}
\bibliography{aaai2026}
\end{small}

\section{Supplementary Materials}

\subsection{Algorithm}
The algorithm of CoDR is presented in Algorithm~\ref{alg:codr}. It consists of two main stages: coreset construction via cluster-pruning (lines 2–9) and dual retrieval from the full support set (lines 10–12).

\begin{algorithm}[htbp]
\caption{CoDR: Coreset-based Dual Retrieval for Demonstration Selection}
\label{alg:codr}
\begin{algorithmic}[1]
\REQUIRE Support set {\boldmath$S$}$= \{x_i,y_i\}_{i=1}^n$, query $x_q$, clustering method $\mathcal{K}$, target coreset size $m$, C-score function $C$, cosine distance $d(\cdot, \cdot)$, shots number $n$
\ENSURE Retrieved demonstration set {\boldmath$S^R$}\\
\COMMENT{Stage 1: Coreset Construction}
\STATE {\boldmath$S^*$} $\leftarrow$ {\boldmath$S$}
\WHILE{$|S^*| > m$}
    \STATE Apply clustering: $K_1, K_2, \ldots, K_k \leftarrow \mathcal{K}$({\boldmath$S^*$})
    \FOR{each cluster $K_i$}
        \STATE Compute intra-cluster distance: $A \leftarrow \mathrm{average}\{d(l, m) \mid l, m \in K_i\}$
        \STATE Compute inter-cluster distance: $B \leftarrow \mathrm{average}\{d(l, m) \mid l \in K_i,\ m \notin K_i\}$

        \STATE Set complexity score: $F_i \leftarrow A \cdot B$
        \STATE Select $\left\lfloor |K_i| \cdot \frac{\exp(F_i)}{\sum_j \exp(F_j)} \right\rfloor$ examples from $K_i$
    \ENDFOR
    \STATE {\boldmath$S^*$} $\leftarrow$ union of selected samples
\ENDWHILE\\
\COMMENT{Stage 2: Dual Retrieval}
\FOR{each sample $s \in \textbf{\boldmath$S^*$}$}
   \STATE Compute weighted C-score: 
   $\hat{C}(s)$ \\
   $\leftarrow \sum_{q^* \in \textbf{\boldmath$S^*$}} [ d(q, q^*) \cdot C(q^*|s) ] \ / \sum_{q^* \in \textbf{\boldmath$S^*$}} d(q, q^*)$
\ENDFOR

\STATE {\boldmath$S^R$} $\leftarrow$ top-$n$ samples in {\boldmath$S$} with highest $\hat{C}(s)$
\RETURN {\boldmath$S^R$}
\end{algorithmic}
\end{algorithm}

\subsection{Additional Results under $8$-shot settings}
The comparison of the performance of $4$-shot and $8$-shot IC, VQA, and FIC on the IDEv2 model are listed in Table~\ref{tab:result_8shotIC}.
Our method (CoDR), demonstrates an advantage in IC and FIC tasks under $8$-shot settings. However, in VQA task, the advantage is minimal, similar to what we observed in $4$-shot settings.
The increase in performance from 4-shot to 8-shot settings is relatively small due to marginal effects. This article does not focus on resolving this issue. Understanding long inputs remains a common challenge for both LLMs and LVLMs, which will be addressed in future work.

\begin{table}[htbp]
\centering
\caption{4-shot and 8-shot IC/VQA/FIC performance with various demonstration selection strategies}
\label{tab:result_8shotIC}
%\scriptsize
\setlength{\tabcolsep}{1mm} 
\begin{tabular}{lcccccc}
\hline
Model-Method & \multicolumn{2}{c}{IC\_CIDEr $\uparrow$} & \multicolumn{2}{c}{VQA\_ACC $\uparrow$} & \multicolumn{2}{c}{FIC\_ACC $\uparrow$} \\
            & 4-shot & 8-shot & 4-shot & 8-shot & 4-shot & 8-shot \\
\hline
{IDEv2}-RG                  & 112.70              & 113.41             & 66.62            & 67.12            & 44.85              & 44.75              \\
{IDEv2}-SITR                & 102.71              & 109.25             & 66.93            & 67.17             & 54.42              & 58.35              \\
{IDEv2}-IRL                 & 111.29              & 112.37             & 67.04       & 67.30
& 50.17              & 55.83              \\
{IDEv2}-CoDR                & 117.54              & 119.20             & 67.01           & 67.31            & 66.28              & 77.85              \\ \hline
\end{tabular}

\end{table}

\subsection{Additional Results on Image Classification}
ImageNet-1k%~\cite{imagenet}
constitutes a widely-used benchmark dataset for image classification tasks, comprising 1,000 distinct object classes. Following the standard splitting protocol, the dataset contains 1,281,167 training images and 50,000 validation images.
The image classification results on ImageNet are shown in Table~\ref{tab:result_imagenet}.
The results show that CoDR achieves state-of-the-art performance. Notably, its advantage is significantly amplified in the IDEv2 model compared to OFv2.
\begin{table}[htbp]
\centering
\caption{1-/2-/4-shot image classification performance with various demonstration selection strategies}
\label{tab:result_imagenet}

\begin{tabular}{lccc}
\hline

Model-Method & 1-shot & 2-shot & 4-shot \\
\hline
OFv2-RG                                           &13.12         & 27.25          & 33.11         \\OFv2-RL                                           & 13.13          & 27.25          & 32.99          \\
OFv2-SITR                                         & 20.59          & 29.84          & 33.02         \\
OFv2-SITRL                                        & 20.73        & 29.96        & 32.79          \\
OFv2-IRL                                          & 21.57         & 28.78        & 33.54         \\
OFv2-CoDR                                         & \textbf{23.50} & \textbf{29.96} & \textbf{33.59} \\ \hline
IDEv2-RG                                          & 12.95          & 36.79      & 43.53         \\
IDEv2-RL                                          & 13.11         & 36.55        & 43.40        \\
IDEv2-SITR                                        & 17.25         & 37.41          & 44.12          \\
IDEv2-SITRL                                       & 20.20          & 37.15          & 42.34         \\
IDEv2-IRL                                         & 13.71         & 36.77         & 41.39 \\
IDEv2-CoDR                                        & \textbf{27.78} & \textbf{43.13} & \textbf{48.22}         \\ \hline
\end{tabular}

\end{table}

\subsection{Additional Results on Random Sampling}
Due to the inherent randomness in the performance of the random strategy, we utilized three different sets of seeds for the random methods (\textit{e.g.}, RG and RL) and K-means (cluster-pruning in CoDR). The results are shown in Table~\ref{tab:result_seed}.
``Mean'' and ``Std'' stand for the average performance and the standard deviation of the $4$-shot ICL, respectively. 

\begin{table}[htbp]
\centering
\caption{4-shot IC/VQA/FIC performance with different random seeds}
\setlength{\tabcolsep}{1mm} 
\label{tab:result_seed}
\begin{tabular}{lcccccc}

\hline
Model-Method & \multicolumn{2}{c}{IC\_CIDEr $\uparrow$} & \multicolumn{2}{c}{VQA\_ACC $\uparrow$} & \multicolumn{2}{c}{FIC\_ACC $\uparrow$} \\
             & Mean & Std & Mean & Std & Mean & Std \\
\hline
IDEv2-RG                  & 112.70               & 0.64              & 66.62               & 0.31              & 44.85               & 0.36              \\
IDEv2-RL                  & 109.30               & 0.61              & 66.93               & 0.34              & 54.42               & 0.33              \\
IDEv2-IRL                 & 111.29               & 1.32              & 67.04               & 0.26              & 50.17               & 0.17              \\
IDEv2-CoDR                & 117.54               & 1.05              & 67.01               & 0.18              & 66.28               & 0.22                     \\ \hline
\end{tabular}
\end{table}

\subsection{Comparison to the learned retrievers}
Recent methods use LVLMs or learned retrievers to select prompts or demonstrations~\cite{feng2024auto,shen2024retrieval}, often requiring a separate model or LVLM invocation per query. While effective, these approaches incur significant computational overhead and depend on large models during retrieval.
Prior works~\cite{yang2023lever,yang2024exploring} have shown that similarity-based retrieval remains a strong baseline in multimodal ICL due to its simplicity and effectiveness. Notably, many learned retrievers still rely on similarity scoring internally.
However, in terms of efficiency, learned retrievers are often slower than even simple similarity-based methods such as SITR, and are substantially outperformed by CoDR in both speed and scalability.
In contrast, CoDR selects demonstrations through a dual retrieval mechanism guided by a compact coreset and a precomputed C-score matrix, avoiding model-heavy retrievers entirely. 
We provide a comparison of several popular learned retrievers in terms of performance and efficiency, as shown in Table~\ref{tab:retriever_comparison}.
LameR~\cite{shen2024retrieval} uses a LVLM as a retriever, resulting in significantly longer time to generate demonstration prompts.
To our knowledge, LeverLM~\cite{yang2023lever} is the only method applicable in multimodal ICL without online query comparison, which shows a strong efficiency. 
In terms of performance, for example, CoDR outperforms LeverLM on the IC task by +2.5 and +3.3 CIDEr on OFv2 and IDEv2, respectively.

\begin{table}[htbp]
\centering
\caption{Comparison to other learned retrievers}
\label{tab:retriever_comparison}
\begin{tabular}{lccccc}
\hline

Model-Method & IC $\uparrow$ &VQA  $\uparrow$ &FIC  $\uparrow$ & Time(s)  $\downarrow$ \\
\hline
OFv2-LameR & 94.24 & 50.30& 36.98     &  23.49 \\
OFv2-LeverLM& 101.68 &51.21&  40.19   &  \textbf{2.93} \\
OFv2-SITR   & 97.29&47.50& 36.98     &  5.37 \\
OFv2-CoDR   &\textbf{104.23} &\textbf{53.49}&\textbf{41.39}& 3.95 \\ \hline
IDEv2-LameR & 105.37 & 62.53&  60.26     &  22.74 \\
IDEv2-LeverLM& 114.23 &66.75&  62.29   & \textbf{ 2.75} \\
IDEv2-SITR  &102.71&66.93 & 54.42    &   4.85  \\
IDEv2-CoDR          & \textbf{117.54}& \textbf{67.01} & \textbf{66.28}    &   3.77  \\ \hline
\end{tabular}

\end{table}

\subsection{Additional Results on Transfer ability}
To evaluate CoDR's transfer capability, we retrieve in-context demonstrations using off-the-shelf OFv2 and IDEv2 models, then perform inference with GPT-4~\cite{achiam2023gpt}, with results in Table~\ref{tab:ta}.
Improvements on classic multimodal tasks like IC and VQA remain modest. We attribute this to GPT-4's likely pretraining exposure to similar datasets, evidenced by its strong $0$-shot performance without image-text demonstrations.
Nevertheless, CoDR consistently outperforms similarity-based and random retrieval.
While in FIC, the results demonstrate the strong transfer ability of CoDR, indicating its effectiveness in selecting demonstrations beyond the models from which they were retrieved.
\begin{table}[htbp]
\caption{Performance on GPT-4 with the retrieval results on OFv2 and IDEv2.}
\label{tab:ta}
\centering
\begin{tabular}{lcccccc}
\hline
Task &0-shot& Rand & Simi& \textbf{OF-CoDR} & \textbf{IDE-CoDR}  \\
\hline
IC  &149.5  & 148.2   &  147.0  &{150.3} & \textbf{150.9}\\
VQA   &\textbf{83.9}& 83.3&82.7& 83.6 & 83.7  \\
FIC  &77.5 & 85.8    &  87.5   & \textbf{89.4}  &88.5 \\
\hline
\end{tabular}%xxxxx
\end{table}

\begin{figure}[htbp]
    \centering
    \includegraphics[width=0.95\columnwidth]{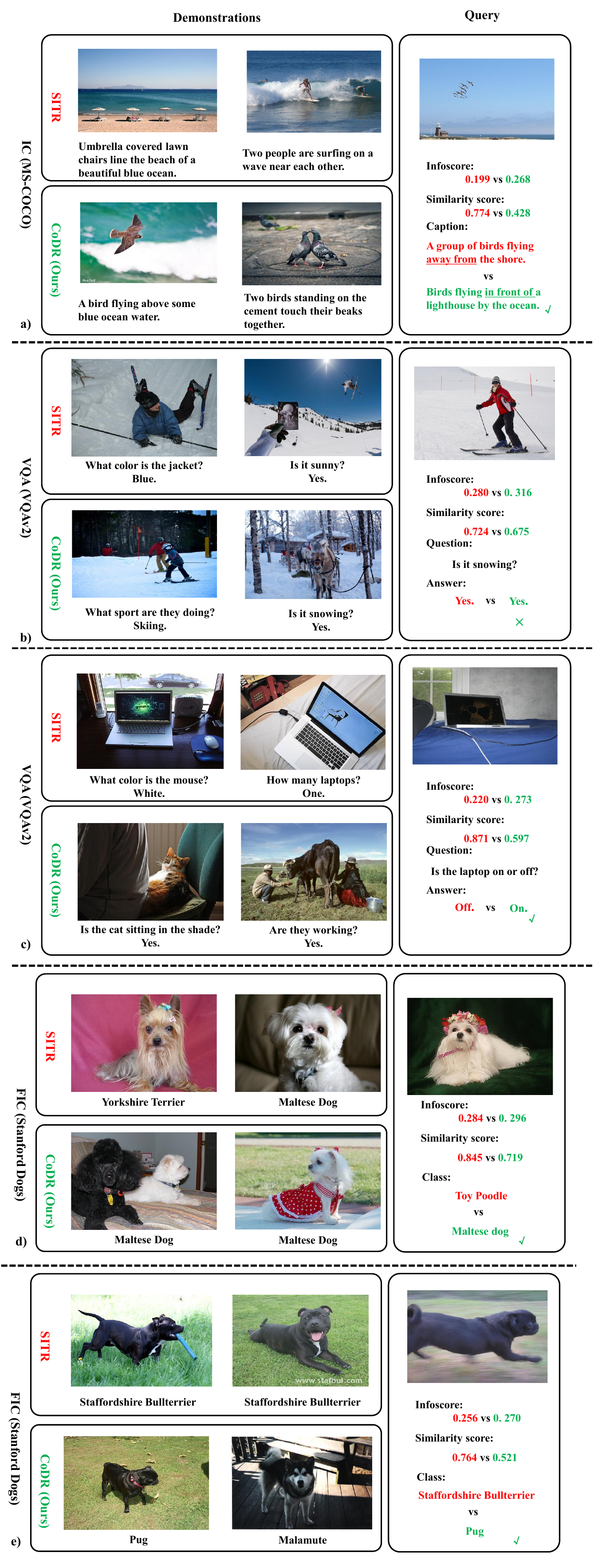}
    \caption{{Visualization of SITR and our method on IC, VQA and FIC tasks.}}
    \label{fig:sup_vis}
\end{figure}

\subsection{Additional Qualitative Results}
We present a qualitative analysis of query-demonstration pairs in Fig.~\ref{fig:vis} to illustrate the advantages of our approach.
Specifically, Fig.\ref{fig:vis}(a) shows an example for IC, (b) and (c) for VQA, and (d) and (e) for FIC.
Among them, Fig.~\ref{fig:vis}(b) illustrates a failure case. When the model encounters a challenging problem, such as determining whether it is snowing from a static image, both CoDR and SITR fail to produce the correct answer.
Additionally, when the query falls outside the distribution represented in the coreset—such as rare visual categories or atypical query structures—CoDR’s similarity-weighted dual retrieval may prioritize semantically distant samples.
However, our CoDR approach still exhibits fewer hallucinations and generates relatively more plausible responses compared to SITR.

\subsection{Prompt Format}
Our prompts remain consistent, as shown in Table~\ref{tab:prompt} for both OFv2 and IDEv2 models.
\begin{table}[htbp]
\caption{Prompt Format of IC, VQA, and FIC}
\label{tab:prompt}
\centering
\begin{tabular}{ll}
\hline
\textbf{Task} & \textbf{Prompt format} \\
\hline
IC   & $\langle$image$\rangle$ Caption: $\langle$Y$\rangle$ \\
VQA  & $\langle$image$\rangle$ Question: $\langle$question$\rangle$ Short answer: $\langle$Y$\rangle$ \\
FIC  & $\langle$image$\rangle$ An image of $\langle$Y$\rangle$ \\
\hline
\end{tabular}
\end{table}

\subsection{Datasets Details}
\subparagraph{MS-COCO~\cite{lin2014microsoft}:}
MS-COCO is widely used in IC, which contains $123,287$ images ($113,278$/$5000$/$5000$ for training/validation/testing in Karpathy split~\cite{karpathy2015deep}). Each image has $5$ corresponding human annotations.
We use the Karpathy training split as the support set and the validation split as the testing queries.

\subparagraph{VQA-v2~\cite{antol2015vqa}:}
VQA-v2 shares the same images with MS-COCO. Each image is associated with multiple open-ended question-answer pairs, resulting in a total of $4,437,570$ pairs. 
We select only the first question-answer pair for each image, ensuring that each image corresponds to a single question-answer pair.
The training and validation splits are consistent with Karpathy split.

\subparagraph{Stanford Dogs~\cite{StanfordDogs}:}
Stanford Dogs is popular in the FIC field, which includes $120$ breeds of dogs, comprising a total of $20,580$ images ($12,000$/$8,580$ for training/testing).
The training and testing splits are used as our testing queries and training set, respectively.

\subsection{Future Works}
Our current CoDR framework focuses on selecting effective demonstrations by leveraging 1-shot performance signals. A natural next step is to explore how to better combine multiple selected demonstrations, potentially incorporating strategies from ensemble learning or prompt optimization.
Furthermore, combining multiple coreset configurations (e.g., from different clustering granularities or random seeds) may offer a more robust and flexible retrieval space for demonstration selection. 

\end{document}